\def\year{2021}\relax
\documentclass[letterpaper]{article} 
\usepackage{aaai21}  
\usepackage{times}  
\usepackage{helvet} 
\usepackage{courier}  
\usepackage[hyphens]{url}  
\usepackage{graphicx} 
\usepackage{natbib}  
\usepackage{caption} 
\usepackage{svg}
\usepackage{multirow}
\usepackage{tabularx}
\usepackage{amssymb}
\usepackage{algorithm}
\usepackage[switch]{lineno}
\usepackage[draft]{hyperref} 
\usepackage[noend]{algorithmic}
\usepackage{etoolbox}\AtBeginEnvironment{algorithmic}{\small}
\usepackage{amsmath}
\title{Interpretable Math Problem Solver via Attributed Grammar}
\author{

}

\title{SMART: A Situation Model for Algebra Story Problems via Attributed Grammar}
\author {
    Yining Hong,
    Qing Li,
    Ran Gong,
    Daniel Ciao,
    Siyuan Huang,
    Song-Chun Zhu\\
}
\affiliations {
    University of California, Los Angeles, USA. \\
    yininghong@cs.ucla.edu,
    \{liqing, nikepupu, danielciao, huangsiyuan\}@ucla.edu,
    sczhu@stat.ucla.edu
}

\begin{document}
\maketitle
\begin{abstract}
Solving algebra story problems remains a challenging task in artificial intelligence, which requires a detailed understanding of real-world situations and a strong mathematical reasoning capability. Previous neural solvers of math word problems directly translate problem texts into equations, lacking an explicit interpretation of the situations, and often fail to handle more sophisticated situations. To address such limits of neural solvers, we introduce the concept of a \emph{situation model}, which originates from psychology studies to represent the mental states of humans in problem-solving, and propose \emph{SMART}, which adopts attributed grammar as the representation of situation models for algebra story problems. Specifically, we first train an information extraction module to extract nodes, attributes, and relations from problem texts and then generate a parse graph based on a pre-defined attributed grammar. An iterative learning strategy is also proposed to improve the performance of SMART further.
To rigorously study this task, we carefully curate a new dataset named \emph{ASP6.6k}.
Experimental results on ASP6.6k show that the proposed model outperforms all previous neural solvers by a large margin while preserving much better interpretability. To test these models' generalization capability, we also design an out-of-distribution (OOD) evaluation, in which problems are more complex than those in the training set. Our model exceeds state-of-the-art models by 17\% in the OOD evaluation, demonstrating its superior generalization ability.
\end{abstract}

\section{Introduction}
\begin{figure}[t]
	\includegraphics[width=\linewidth]{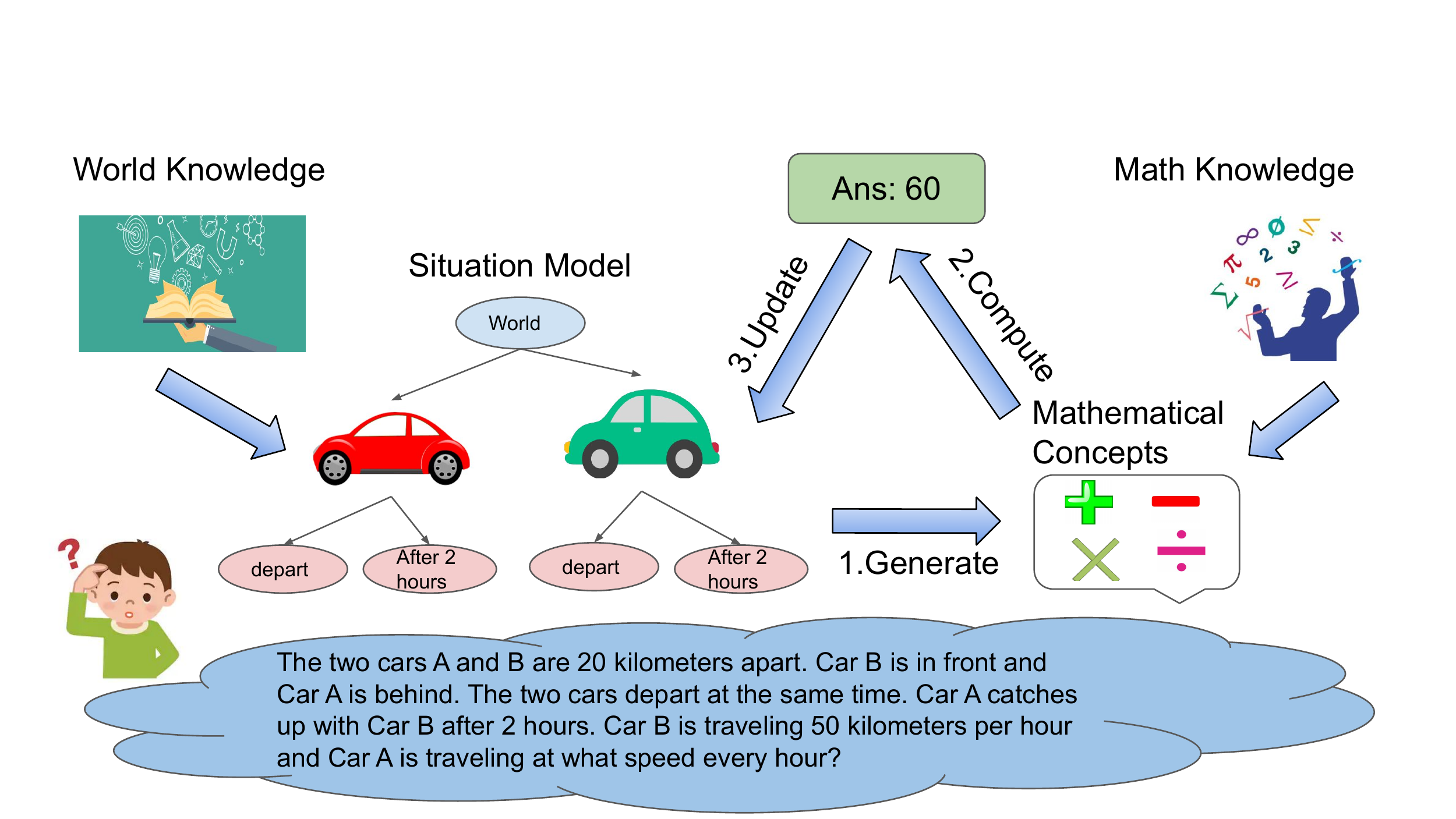}
	\caption{The process of human solving algebra story problems: We first hallucinate a situation model from the text and then perform arithmetic reasoning on the situation model to compute an answer. If we fail to generate a correct solution, we can adjust our situation model accordingly.}
	\label{fig:teaser_aog}
\end{figure}

\textit{Algebra Story Problems}, depicted by~\citet{hinsley1977} as ``twentieth-century fables'', remain a critical challenge in modern artificial intelligence. An algebra story problem typically describes a real-world situation and inquires about an unknown fact in the situation. It goes beyond arithmetic since one has to first comprehend the situation, recognize the goal in the problem, and then develop a solution for it ~\cite{nathan1992theory}. Psychology
 studies~\cite{bjork2013cognitive,abedi2001language} also indicate that algebra story problems can serve as a test of children's cognitive skills to perform arithmetic reasoning on real-world tasks. 
However, although algebra story problems are distinguished \textit{per se}, related works from the community of artificial intelligence and natural language processing often mix them with other types of problems, such as number problems and geometry problems, into one whole task called \textit{Math Word Problems} (MWPs)~\cite{wang-etal-2017-deep, Huang2016HowWD, Amini2019MathQATI}

Recent works on Math Word Problems~\cite{wang-etal-2017-deep, Huang2018NeuralMW, Wang2018TranslatingMW, Xie2019AGT, hong2021lbf} focused on using end-to-end neural networks (\textit{e.g.}, Seq2Seq, Seq2Tree) to directly translate a problem text into an expression, which is then executed to get the final answer. 
Although they seem to obtain satisfying performance, such end-to-end neural models suffer from the following drawbacks:
\begin{itemize}
    \item Lack of interpretability. The expressions generated by neural networks are hard to interpret without the intermediate problem-solving process. An exemplary expression from~\autoref{fig:qualitative} is ``(24+60)/[1-(1-(2/5))*(3/10)-(2/5)*(3/4)-(2/5)]", which makes no sense to humans, even though it generates the correct answer. 
    \item Lack of generalization ability. These neural solvers usually fail in scenarios that are more sophisticated than those in training.
\end{itemize}

To address these issues in current research on Math Word Problems, we make the following efforts in this work. 

First, we curate a new benchmark named \textit{ASP6.6k}, which contains four canonical types of algebra story problems: motion, price, relation, and task. We build our dataset upon Math23K~\cite{Wang2017DeepNS}, the most frequently-used MWP dataset in recent years, and categorize algebra story problems by following precisely the criteria set by \citet{mayer1981frequency}.

Second, we introduce the cognitive concept of a \textit{situation model}~\cite{dijk1983}, which is widely used in psychology studies to model the mental states of humans in problem-solving~\cite{Reusser1990FromTT, greeno1989, nathan1990, CoquinViennot2007ArithmeticPA, Leiss2010TheRO}.
It is believed that problem-solving techniques, such as mathematics and logic, are applied to the hallucinated situation model instead of the problem text. As shown in \autoref{fig:teaser_aog}, the situation model interacts with mathematical concepts to derive a solution for the problem.

To efficiently represent the situation model, we propose \textit{SMART}, which utilizes attributed grammar \cite{knuth1990genesis, Liu2014SingleView3S, Park2015AttributedGF, Park2016AttributeAG} as the representation of a situation model in algebra story problems. More specifically, the world, agents, and events depicted in an algebra story problem are represented as nodes in a hierarchical parse graph, derived from the problem text with a context-free grammar. The parse graph nodes are further augmented with attributes to represent quantities in the problem, and the relations between these quantities are encoded as numerical constraints on the corresponding attributes. The construction of these constraints usually requires both commonsense knowledge and mathematical knowledge. Therefore, the parse graph generated by an attributed grammar can capture the situation model's desired characteristics for algebra story problems. With the parse graph, the problem solving is equivalent to seeking one unknown attribute in the graph and can be formulated as an attribute propagation process guided by the constraints on these attributes. To automatically construct parse graphs for problems, we first train an information extraction module to extract nodes, attributes, and relations from problem texts and then generate parse graphs based on a carefully designed attributed grammar.


The learning of SMART is nontrivial. Since the grammar parsing and the problem solving are non-differentiable, we cannot use back-propagation to learn SMART in an end-to-end fashion under the supervision of final answers. Therefore, we propose a two-stage learning strategy: First, we manually design a text parser to generate initial supervision on parse graphs and use them to train the information extraction module in SMART. Second, we adopt an iterative learning method to strength the information extraction module, where pseudo-gold parse graphs at each iteration augment the supervision for the next learning iteration.


We conduct experiments on the newly curated benchmark ASP6.6k, and the proposed SMART model outperforms all neural network baselines by a large margin. Moreover, it demonstrates stronger generalization ability in an out-of-distribution evaluation, where the test problems are more complex than those in the training set. A qualitative study also suggests that SMART achieves better interpretability and generalization ability than the neural models.

\section{Related Works}
\paragraph{Math Word Problems}
Solving math word problems has attracted researchers for decades. 
Early solvers \cite{Fletcher1985UnderstandingAS, Bakman2007RobustUO, Yuhui2010FrameBasedCO} use rule-based methods which are generally fixed and only work on single-step word problems for one category of problems. The next stage of solvers use semantic parsing techniques \cite{Hosseini2014LearningTS, KoncelKedziorski2015ParsingAW, Shi2015AutomaticallySN, Huang2017LearningFE}. These methods attempt to parse the problem text into an intermediate structured representation, usually annotated in the training set. Specifically, \citet{Shi2015AutomaticallySN} uses Context Free Grammar to solve number problems, which is quite different from our grammar for story problems. Statistical learning methods~\cite{Kushman2014LearningTA, Zhou2015LearnTS, Mitra2016LearningTU,  Roy2017UnitDG, Huang2016HowWD} attempt to boost semantic parsing techniques, like choosing the most probable template to use \cite{Mitra2016LearningTU}. However, these templates are still fixed before training, leading to inflexibility in solving more sophisticated problems. 

Researchers recently focus on solving math word problems using neural networks  \cite{Ling2017ProgramIB, wang-etal-2017-deep, Huang2018NeuralMW, Robaidek2018DataDrivenMF, Wang2018TranslatingMW, Wang_Zhang_Zhang_Xu_Gao_Dai_Shen_2019, Chiang2019SemanticallyAlignedEG, Xie2019AGT, zhang2020graph2tree, hong2021lbf}.  The mere translation from a text to an equation neglects the intermediate process required by problem solving, thus lacking interpretability. In this paper, we seek to combine the strengths of both symbolic reasoning and neural networks, where we use neural modules to update symbolic representations.

\paragraph{Situation Model}
Situation models have been proven crucial in human discourse comprehension and problem solving~\cite{zwaan1995dimensions,nesher2003situation}. Researchers have long believed that text comprehension is a process of construction and integration~\cite{gernsbacher2013language, kintsch1998comprehension}. \citet{hegarty1995comprehension} indicate that without a situation model, problem solvers with a direct translation approach are more likely to fail for math problems. A recent study \cite{raudszus2019situation} also shows that the ability of building a situation model is a strong indicator of cognitive and linguistic skills. 

There is a history of situation model construction for algebra story problem solving~\cite{Reusser1990FromTT, greeno1989, nathan1990, CoquinViennot2007ArithmeticPA, Leiss2010TheRO}. \citet{kintsch1985understanding} use a situation model to analyze processing requirements and difficulties of algebra word problems. \citet{nathan1992theory} build a situation model to predict student mental state and predict how to tutor students based on interaction history. In contrast, this paper builds a situation model for the machine, which uses attributed grammar to model the problem solving for algebra story problems.
A situation model should satisfy the following properties \cite{dijk1983} : 
\begin{itemize}
    \item Reference: The situation model should represent the world the text is stating about.
    \item Coherence: All facts, implicit or explicit, need to be connected as long as the relations are indicated by the text.
    \item Situational Parameters: It includes the parameters and attributes about the world and events in the text.
    \item Event Independence: It needs to be invariant regardless of the number of events and their order. 
\end{itemize}
We argue that a parse graph derived from attributed grammar can capture the above properties of a situation model.

\paragraph{Attributed Grammar}
Attributed grammar is proposed by Knuth to handle the semantics of programming languages~\cite{knuth1990genesis}. In recent years, researchers use attributed grammar to represent hierarchical grammar structures for images \cite{han2005bottom,wang2013weakly}, video events~\cite{lin2009semantic}, human poses~\cite{park2016attribute}, indoor scene understanding~\cite{qi2018human,jiang2018configurable,huang2018holistic,chen2019holistic++}, \textit{etc}. The attributes are assigned to terminal and non-terminal nodes of a grammar based on commonsense knowledge. Attributes between terminal nodes and non-terminal nodes are related by soft constraints or hard constraints, depending on the specific task. We utilize hard constraints in SMART for mathematical reasoning. In addition, attributes in a parse graph can be propagated in a controlled and formal way. 
\\


\section{The \textit{ASP6.6k} Dataset}
We curate the new dataset \textit{ASP6.6k} from the widely used Math23K dataset. To select and categorize algebra story problems, we first compute the term frequency–inverse document frequency (TF-IDF) features for each problem in the dataset and then use k-means clustering to group problems into different categories. We use the elbow method \cite{Thorndike1953WhoBI} to find the optimal K for the clustering. To further remove noise, we manually select certain keywords to filter out problems that do not belong to the group. 
We select a subset of problems from Math23K following the criteria from \cite{mayer1981frequency,nathan1992theory}:
\begin{itemize}
    \item The problem ask for numerical answers rather than translating a story into equations.
    \item The problem has a story-line consisting of characters, objects, and/or actions. 
\end{itemize}
As a result, we obtain a dataset of 6666 problems spanning from four typical types of algebra story problems: motion, price, relation, and task. 

Here, we provide a brief summary of the problem types:
\begin{itemize}
    \item \textbf{Motion}: problems involve traveling and require understanding of per time rate.
    \item \textbf{Task}: problems involve completion of tasks and require understanding of the relations between fractions.
    \item \textbf{Price}: problems involve purchasing items and require understanding of unit price and total price.
    \item \textbf{Relation}: problems involve a description of relationship between two objects.
\end{itemize}

\begin{table}[htbp]
\centering
\small
\begin{tabular}{ |c|c|c|c|c|c| }
\hline
  & \textbf{Motion} & \textbf{Task} & \textbf{Relation} & \textbf{Price}& \textbf{Total} \\ 
 \hline
 Problems & 1687 &1158 &1915 & 1908  & 6666\\  
 \hline
 Avg. Length  & 37.0 & 33.6 & 28.8 & 26.7 & 31.1\\
 \hline
 Avg.Agents & 1.85& 1.13& 2.34& 1.96& 1.90\\
 \hline  
 Avg. Events & 1.93& 2.21& 2.58& 2.30& 2.27\\
 \hline
 Avg. Relations & 3.07& 3.32& 3.53& 2.98& 3.22\\
 \hline
\end{tabular}
\caption{Dataset Statistics. Length is number of tokens.}
\label{tab:stats}
\end{table}

\begin{table*}[t]
\centering
\small
\begin{tabularx}{\textwidth}{|l|X|}
\hline
  Problem Type & Sample Problem\\ 
 \hline
 Task  & The engineering team built a viewing trail and completed 30\% of the full length in the first week and 45\% of the full length in the second week. 150 meters in two weeks, how long is the length of this trail?\\  
 \hline
 Motion & Mingming’s family went to travel, they took a 14-hour train ride, and then a 5-hour car ride before reaching their destination. It is known that the speed of the train is 120 kilometers/hour and the speed of the car is 60 kilometers/hour. How long is this journey?\\
 \hline
 Relation & Xiaogang's weight is 28.4 kg, Xiaoqiang's weight is 1.4 times that of Xiaogang, Xiaoqiang's weight = how many kilograms? \\
 \hline
 Price & The school bought 45 sets of desks and chairs at 128 yuan per desk and 52 yuan per chair. How much did it spend?\\
 \hline
\end{tabularx}
\caption{An example of each problem type.}
\label{tab:sample}
\end{table*}

Details of the dataset statistics are listed in Table \ref{tab:stats}.
Examples for each problem type are shown in Table \ref{tab:sample}. See supplementary materials for more details about dataset preprocessing and more statistics.



\section{SMART}
\begin{figure*}[t]
    \centering
    \includegraphics[width=\textwidth]{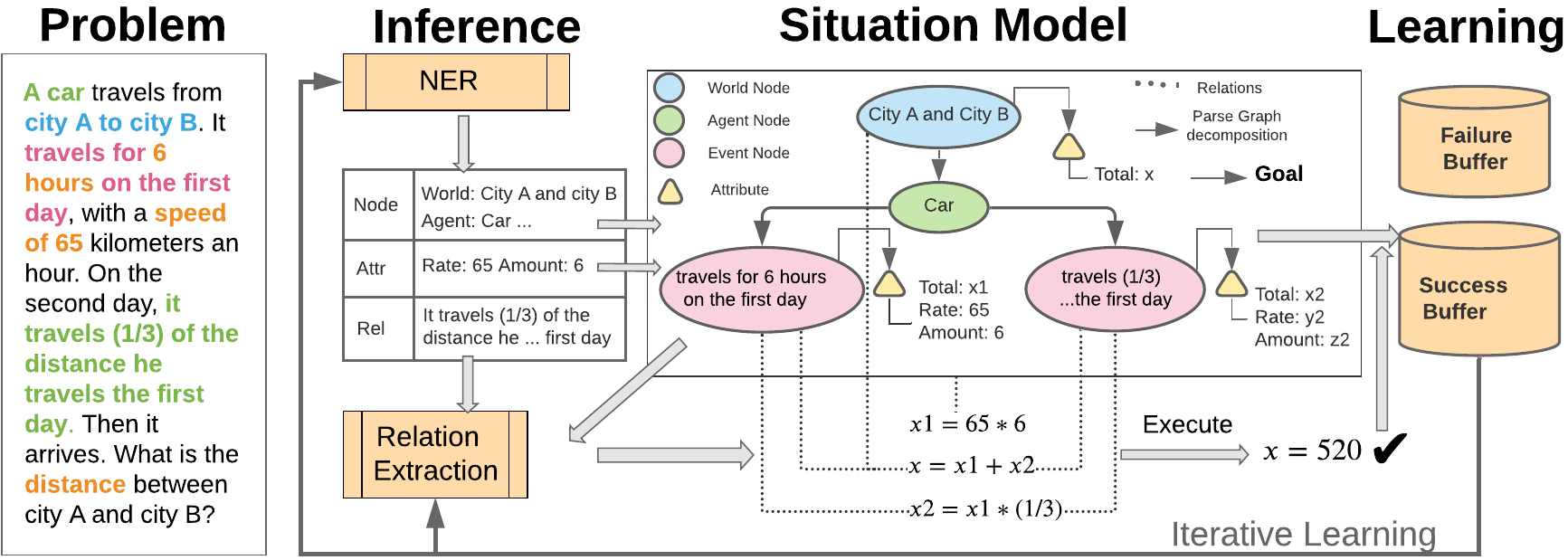}
    \caption{Overview of our SMART model. The Named Entity Recognition (NER) module extracts the spans of nodes, attributes, as well as relations from the text, and construct a parse graph using Attributed Grammar. The Relation Extraction module uses the relation spans and the parse graph already constructed to embed some relations into the parse graph. In the updated graph parser, Relation Extraction corresponds to Seq2Seq. The relations are then executed to get the final answer. If the answer is correct, it is added to the buffer of pseudo-gold parse graphs to train NER and Seq2Seq. If not, it is added to the failure set to be updated in the following iterations.}
    \label{fig:framework}
\end{figure*}
In this section, we introduce SMART, a situation model for algebra story problems via attributed grammar.

\subsection{Attributed Grammar}

Inspired by \citet{qi2018human}, an attributed grammar is designed for the domain of algebra story problems, as shown in \autoref{tab_attributed_grammar}.
\begin{table}[H]
\centering
\begin{tabular}{l}
\hline
$G = (S, V, A, E, R)$ \\
\hline
$S$ is the start symbol. \\
$V$ = \{S, World, Agents, Agent, Events, Event\} \\
$A$ = \{rate, amount, total\} \\
$E$ = \{$e$: $e$ is a valid equation on attributes.\}\\
$R$ = \{$S$ $\to$ World\\
~~~~~~~~World $\to$ Agents\\
~~~~~~~~Agents $\to$ Agents Agent $|$ Agent\\
~~~~~~~~Agent $\to$ Events\\
~~~~~~~~Events $\to$ Events Event $|$ Event\}\\
\hline
\end{tabular}
\caption{The attributed grammar for algebra story problems.} \label{tab_attributed_grammar}
\end{table}

In attributed grammar, the production rules $R$ are designed by the following observation: a problem usually depicts a \textit{world}, where several \textit{agents} perform several \textit{events}.

Inspired by \cite{Roy2017UnitDG,nathan1992theory,mayer1981frequency},  we design three types of attributes to augment the nodes: i) \textit{rate}: a quantity which is certain measure corresponding to one unit of some other quantity, indicated by phrases like ``A per B" and ``each A has B" (\textit{e.g.}, speed, price). ii) \textit{amount}: a measurement of units of rate quantities (\textit{e.g.}, hour). iii) \textit{total}: a quantity which equals to the multiplication of rate and amount (\textit{e.g.}, distance).

The relations $E$ represent possible constraints on the attributes in the form of equations. These constraints can be either explicitly stated in the text, such as ``it travels 1/3 of the distance'', or implied by commonsense knowledge, such as ``distance = speed $\times$ time''.
See \autoref{fig:framework} for an exemplary parse graph generated from the attributed grammar.

\subsection{Grammar Parsing}
The construction of the situation model for an algebra story problem is equivalent to parsing the problem text into a parse graph. Formally, given the problem $x$ and the attributed grammar $G$, the parsing process is formulated as:
\begin{equation}
pg^{\ast} = \arg\max_{pg \in \mathcal{L}(G)} p(pg \mid x),
\end{equation}
where $\mathcal{L}(G)$ denotes the language of the attributed grammar. The probability of a parse graph $pg$ given $x$ can be written as a joint probability of its nodes $V_{pg}$, attributes $A_{pg}$ and relations $E_{pg}$:
\begin{align}
p(pg \mid x) &= p(V_{pg}, A_{pg}, E_{pg}|x) \\
&= p(V_{pg}|x) \cdot p(A_{pg}|x) \cdot p(E_{pg}|x)
\end{align}
Here we assume the independence of these nodes, attributes, and relations to simplify our model and leave the exploration of their dependency for future works.

The extraction of nodes, attributes and relations is achieved by a three-step process. 

First, we define seven categories of entities: nodes (\textsc{World, Agent, Event}), attributes (\textsc{Rate, Amount, Total}), \textsc{Rel} (which denotes a text span that indicates relation). We train a named entity recognition (NER) system to recognize these entities from the text. Specifically, we have one NER model to extract the attributes, and another one for the extraction of the nodes and \textsc{Rel}. We use Nested NER \cite{Strakov2019NeuralAF} for the second model. We use BERT-chinese-base pre-trained model and fine-tune it on our NER task. We then have:
\begin{align}
p(V_{pg}|x) &= \frac{1}{|V_{pg}|} \sum_{w \in V_{pg}} p_{ner}(w) \\
p(A_{pg}|x) &= \frac{1}{|A_{pg}|} \sum_{w \in A_{pg}} p_{ner}(w)
\end{align}
where $|V_{pg}|$ is the length of a node span, $|A_{pg}|$ is the length of an attribute span, and $w$ is a word in the node or attribute span. $p_{ner}(w)$ is the probability of a word being labelled as a specific category.

Second, we connect these nodes and attributes into a parse graph based on two distances: the word distance between two nodes in the problem text, and the distance (number of links) between them in the dependency parse. Some constraints in the dependency parsing are also imposed, \textit{e.g.}, the noun word representing an agent is preferred to be the \textsc{nsubj} of the verb word in the event node.
Please refer to the supplementary materials for the complete list of rules and constraints used in the parsing. Attributes not extracted by NER are marked as an unknown.

Third, we train a Seq2Seq model to translate the \textsc{Rel} entity from a text span into an equation $e$, which consists of attributes in the parse graph and arithmetic operators ($\{+,-,\times,\div, \wedge, =\}$). To include attributes in the equation, the input to the Seq2Seq model is the concatenation of \textsc{Rel}, nodes and the attributes of each node. See supplementary for the examples of inputs and outputs of the Seq2Seq model. The probability of the relation is defined as:
\begin{equation}
    p(E_{pg}|x) = \sum_{e \in E_{pg}} p_{seq2seq}(e|\text{\textsc{Rel}},V_{pg}, A_{pg})
\end{equation}
where $p_{seq2seq}$ is the Seq2Seq output probability.

\subsection{Goal Recognition}
The goal of a problem is usually an interrogative word extracted by NER (\textit{e.g., how many}). It can be in an attribute or in \textsc{Rel}. We transform the interrogative word into an unknown in the equation.

\subsection{Problem Solving}
 After building the parse graph, the problem can be easily solved by feeding the relation equations and the goal into a mathematical solver~\cite{10.7717/peerj-cs.103}. This stage requires mathematical reasoning skills, which are perfectly provided by a stand-alone solver.

\section{The Learning of SMART}
The learning objective of SMART is to optimize the information extraction module including both the NER system and Seq2Seq models for relation and goal. Since the grammar parsing and the problem solving are non-differentiable, we cannot use back-propagation to learn SMART in an end-to-end fashion under the supervision of final answers. Therefore, we propose a two-stage learning strategy: First, we manually design a text parser to generate initial supervision on parse graphs and use them to train the information extraction module in SMART. Second, we adopt an iterative learning method to strengthen the information extraction module.

\subsection{Initial Supervision}
Since we do not have the ground-truth parse graphs, we generate parse graph proposals using a hand-designed text parser. The parser extracts nodes, attributes and relations from the text to construct a parse graph. 

\paragraph{Attribute Extraction}
We first extract all numbers of  \textit{rate}, \textit{amount} and \textit{total} using pos Tagger by spaCy\footnotemark. 
\footnotetext{https://spacy.io/}
Similar to \citet{Roy2017UnitDG}, we refer to the unit of attribute \textit{total} to be ``Num Unit" (short for Numerator Unit), and the unit of attribute \textit{amount} to be ``Den Unit" (short for Denominator Unit). The unit of attribute \textit{rate} is therefore ``Num Unit per Den Unit". \autoref{tab:ruq} shows the attributes and attribute units of an exemplar problem. 

\begin{table}[htbp]
    \centering
    \small
    \resizebox{\linewidth}{!}{
    \begin{tabular}{l|c|c|c}
        \hline
        \multirow{4}{*}{\shortstack{\textbf{Problem}: Each kilogram of pears\\ cost 3.65 dollars. How many\\ dollars does mom have to\\ pay for 13 kilograms of pears?}} &\textbf{\textit{rate}} & \textbf{\textit{amount}} & \textbf{\textit{total}}\\
        \cline{2-4}
        &3.65 & 13 & how many\\
        \cline{2-4}
        &\textbf{Num Unit} & \textbf{Den Unit} &\\
        \cline{2-3}
         & dollar & kilometer&\\
        \hline
    \end{tabular}}
    \caption{The attributes (\textit{rate}, \textit{amount}, \textit{total}), Num Unit, Den Unit of an exemplar problem}
    \label{tab:ruq}
\end{table}
Generally, when we have a word marked as ``NUM" or ``X" (in the case of fractions) by pos tagger, or when the word is ``how many" (in the case of  interrogative phrase), we check if there's word such as ``per" or ``each" nearby. If so, we mark the number as \textit{rate} and extract the Num Unit and Den Unit. We then extract \textit{amount} and \textit{total} which are followed by Den Unit and Num Unit respectively.  

\subsubsection{Node Extraction}
The next step is to extract the nodes and link attributes to nodes, based on the POS tagging and the dependency parsing. Specifically, we extracts the nouns in the text and treat them as agent nodes. We extract the verbs and their dependents as event nodes. The world node represents the scope of a problem, and is associated with a \textit{total} attribute denoting the total quantities to be covered in the problem. The extraction is conditioned on the problem type. For the type of motion, it means the total distance within the scope of the problem; for the type of price, it is the total money that can be spent; for the type of task completion, it is usually the total amount of tasks to be completed. The world node is extracted based on rules using POS tagger, dependency parser and regular expressions. If there is no ``scope" information in the problem, we just place a default world node in the parse graph.  

We connect nodes and attributes based on distance and dependency parsing.

\paragraph{Relation Extraction}
We represent the relations explicitly stated in the problem text via the first-order logic:
\begin{itemize}
    \item \textbf{Variables:} We define a node $v$ to be a variable.
    \item \textbf{Functions:} We consider an attribute to be a function of a node, \textit{i.e.}, Rate($v$), Amount($v$), Total($v$). Moreover, we define two extra functions: a sum function which takes in the same attribute of several nodes and returns their sum, e.g., {Sum(Total($v_i$), Total($v_j$))};
    a left function, which computes the quantities that haven't been covered by events so far, e.g., {Left(Total($S$), Total($v_i$), Total($v_j$))}. 
    \item \textbf{Predicates:} We view relations as predicates. Predicates take in functions $F$, and sometimes a value $n$ representing numerical relation (if $n$ is detected by relation extraction, we exclude it from the attribute set). These include:  {Equal}($F(v_i)$, $F(v_j)$); {More\_than($F(v_i)$, $F(v_j)$, $n$)}; {Less\_than($F(v_i)$, $F(v_j)$, $n$)}; {Times\_of($F(v_i)$, $F(v_j)$, $n$)}.  
\end{itemize}
Please refer to the supplementary materials for the complete definitions for functions and predicates.

To mine relations from the text, we first use keyword matching to measure how much a text span is considered to indicate a relation based on keywords (e.g., more than, less than, equals to, times of, left) and then represent the relation as a first-order logic predicate. The predicates are transformed into equations.





\subsection{Iterative Learning}
To further improve the performance of SMART, we propose an iterative learning method: we keep a success buffer, which stores the pseudo gold parse graphs generating correct answers, and a failure buffer, which keeps track of the problems not being solved yet. The pseudo gold parse graphs provided by initial supervision served as the initialization of the success buffer for the first iteration. At each iteration, we first use the success buffer to update the model, and then apply the updated model to the instances in the failure buffer to check if the updated model can solve new problems. The details of the iterative learning method are illustrated in Algorithm~\autoref{alg:iterative}. 

\begin{algorithm} [H]
\begin{algorithmic}[1]
\caption{Iterative Learning}\label{alg:iterative}
\STATE \textbf{Input}: training set $\mathcal{D}=\{(x_i, y_i)\}_{i=1}^N$
\STATE Success buffer $\mathcal{B}$, Failure buffer $\mathcal{F}$, updated parser $\theta$
\STATE \hfill$\triangleright$\textit{Parse Graph Proposal}
\FOR {$x_i, y_i\in \mathcal{D}$} 
    \STATE ${pg}_i$ = initial\_parser ($x_i$)
    \IF {execute(${pg}_i$) = $y_i$ }
        \STATE $\mathcal{B} \leftarrow \mathcal{B} \cup \{x_i, pg_i\}$
    \ELSE
        \STATE $\mathcal{F} \leftarrow \mathcal{F} \cup \{x_i, y_i\}$
    \ENDIF
\ENDFOR

\STATE \hfill$\triangleright$\textit{Iterative Learning}
\WHILE {not converge}
    \FOR {$x_i, pg_i\in \mathcal{B}$}
         \STATE $\theta = \theta - \nabla_{\theta} J(x_i, pg_i)$
    \ENDFOR
    \FOR {$x_i, y_i\in \mathcal{F}$}
         \STATE ${pg}_i$ = updated\_parser ($x_i$)
        \IF {execute(${pg}_i$) = $y_i$ }
            \STATE $\mathcal{B} \leftarrow \mathcal{B} \cup \{x_i, pg_i\}$
            \STATE remove $\{x_i, y_i\}$ from $\mathcal{F}$
        \ENDIF
    \ENDFOR
\ENDWHILE

 \end{algorithmic}

 \end{algorithm}
\section{Experiments}
\begin{figure*}
    \centering
    \includegraphics[width=\textwidth]{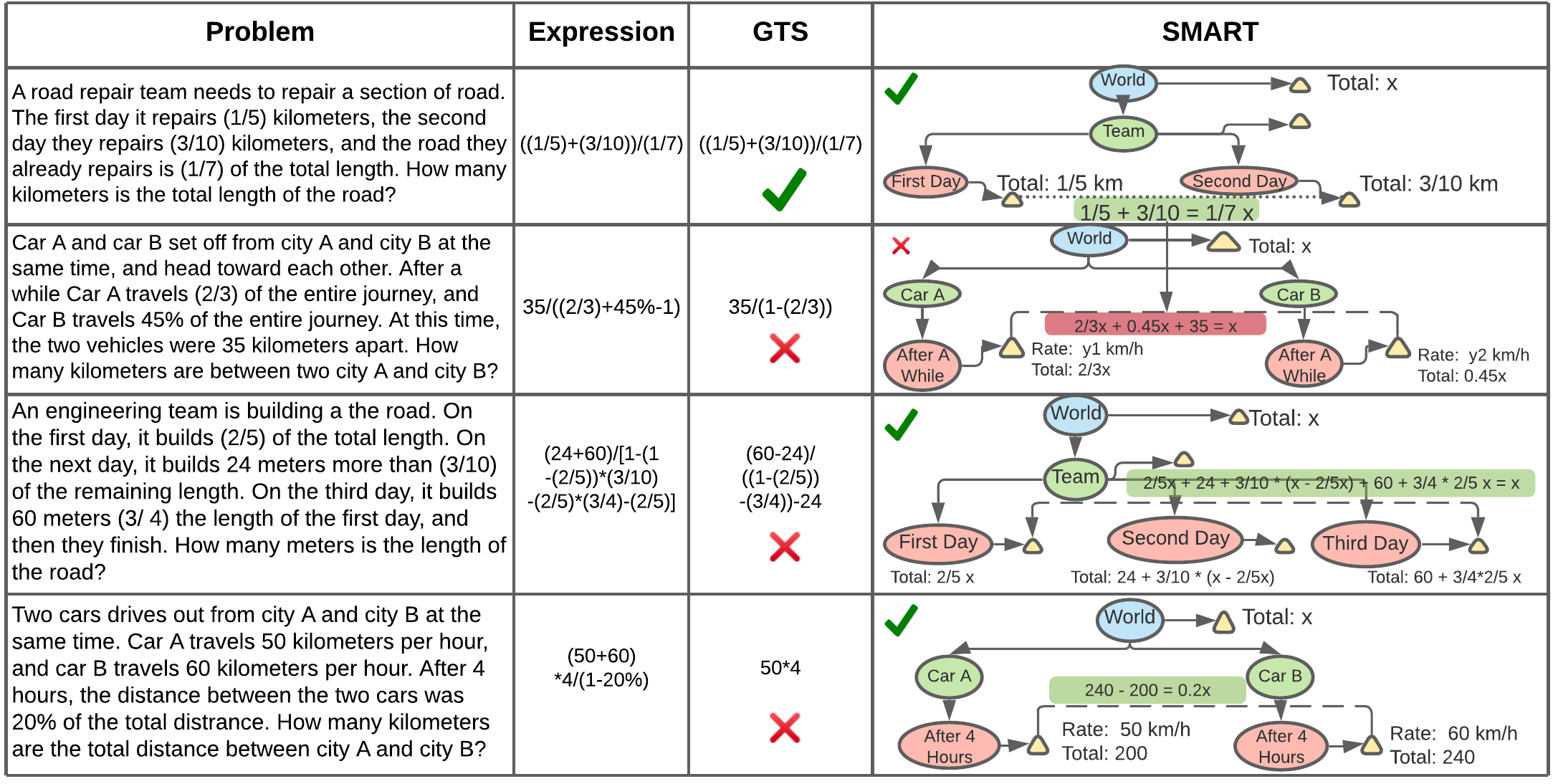}
    \caption{Qualitative study of the GTS model and our SMART model. For the first two instances, we visualize the result in the original test set. The last two instances are from the OOD dataset, where the test set has greater problem length in general compared to the training set.}
    \label{fig:qualitative}
\end{figure*}

\subsection{Experimental Setup}
\paragraph{Dataset} We evaluate our SMART model on the newly curated  ASP6.6k dataset. The final dataset is randomly divided into
training and test sets of 5,332 and 1,334 problems (i.e, approximately a 80/20 split).

\paragraph{Evaluation Metric}
We report the answer accuracy of the models: the generated solution is considered correct if it executes to the ground-truth answer. 
Furthermore, we design an out-of-distribution (OOD) evaluation to examine the models' generalization ability.

\paragraph{Baselines}
We compare the proposed SMART model with several state-of-the-art neural models for math word problems:  MathEN~\cite{Wang2018TranslatingMW}, Group-ATT~\cite{Wang_Zhang_Zhang_Xu_Gao_Dai_Shen_2019}, GTS\cite{Xie2019AGT}, and Graph2Tree \cite{zhang2020graph2tree}.

\subsection{Results and Analyses}
\paragraph{Comparison with State-of-the-art Models}
\autoref{tab:acc} summarizes the comparison of the answer accuracy on the test set with regard to problem types. The proposed SMART model significantly outperforms all the neural models and beats the state-of-the-art model by nearly 3\% in terms of the overall accuracy. More specifically, SMART outperforms the neural models by 3\% and 4\% on the Motion and Relation problems, while it achieves comparable performance on the Task problems and lower performance on the Price problems. 
\begin{table}[htbp]
    \centering
    \small
    \begin{tabular}{c|c|cccc}
    \hline
    \textbf{Model}  &  \textbf{Overall} & \textbf{Motion} & \textbf{Task} & \textbf{Relation} & \textbf{Price}\\
    \hline
    MathEN     & 67.8 &68.3&70.2&63.3&70.5 \\
    GroupATT  & 67.4 &65.2&70.7&63.6&71.5\\
    GTS & 76.8 & 73.2 & 72.1 & 76.0 & \textbf{83.6} \\
    Graph2Tree  & 76.8 &76.9&\textbf{79.0}&73.8&78.7\\
    \hline
    SMART &  \textbf{79.5} &\textbf{79.8}&\textbf{79.0}&\textbf{77.9}&81.8\\
    \hline
    \end{tabular}
    \caption{The answer accuracy on the test set (\%). 
    }
    \label{tab:acc}
\end{table}

\begin{table}[htbp]
    \centering
    \small
    \begin{tabular}{c|c|cccc}
    \hline
    \textbf{Model}  &\textbf{Overall} & \textbf{Motion} & \textbf{Task} & \textbf{Relation} & \textbf{Price}\\
    \hline
    MathEN     & 31.7 & 22.6&28.9&39.9&33.2\\
    GroupATT  &35.0 &24.0&42.2&42.6&32.7\\
    GTS & 45.8 &44.5 &41.9 &49.9 &45.3\\
    Graph2Tree  & 45.1 & 34.1 & 47.4 & 55.1 & 41.9\\
    \hline
    SMART & \textbf{63.2} &\textbf{65.0} &\textbf{64.8} &\textbf{62.9} &\textbf{60.8}\\
    \hline
    \end{tabular}
    \caption{The answer accuracy in the OOD evaluation (\%). The test set is the 20\% longest problems of each type.}
    \label{tab:ood}
\end{table}

\paragraph{Out-of-distribution Evaluation}
To measure the models' generalization ability, we conduct an out-of-distribution (OOD) evaluation, where the test set contains more complex problems than the training set. The length of an algebra story problem is a good proxy for its solving complexity.
Therefore, we select the longest 20\% of problems for each type as the test set and the rest as the training set. 
\autoref{tab:ood} summarize the answer accuracy of all models in the OOD evaluation. The results show that SMART has better generalization ability. It outperforms the neural models by 17\%, revealing SMART's strong ability to reason about more complicated situations even if it is trained on much simpler problems.

\begin{figure}[htbp]
    \centering
    \includegraphics[width=\linewidth]{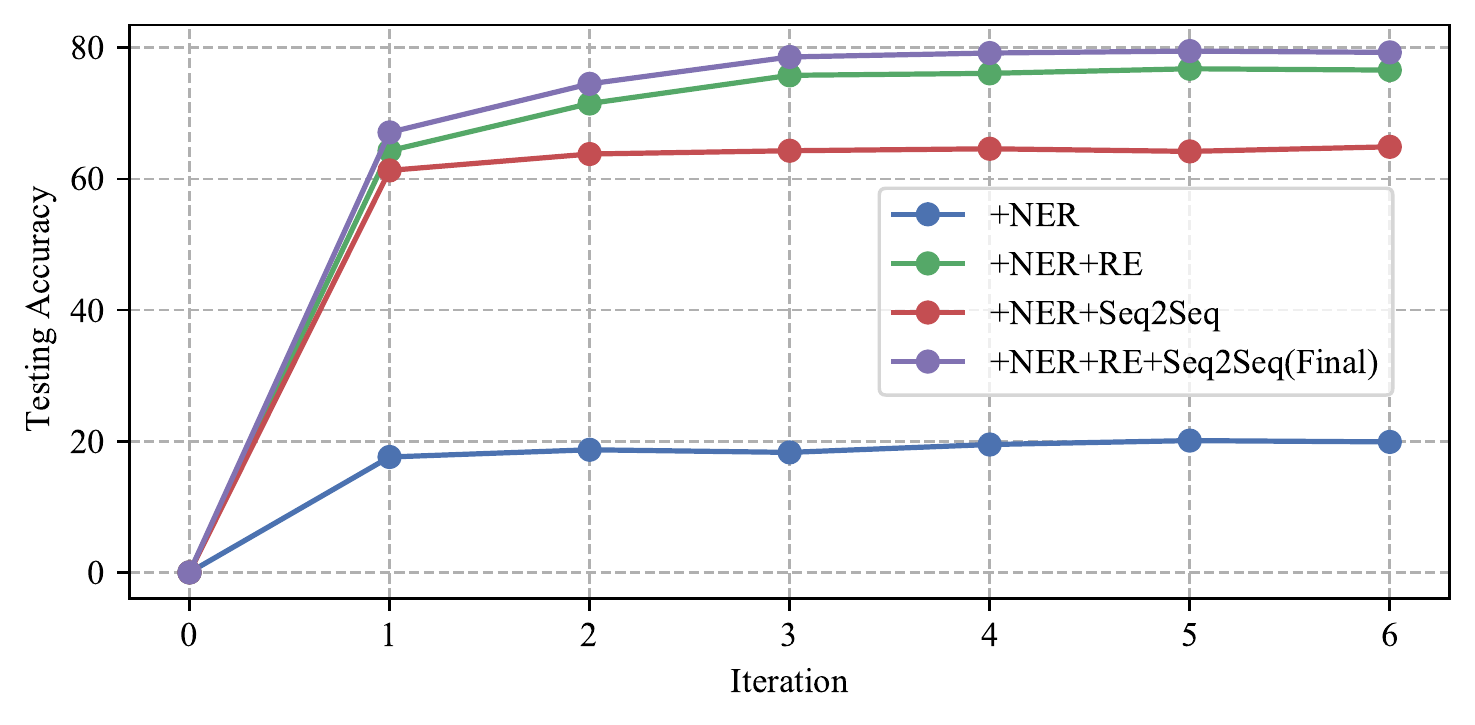}
    \caption{The answer accuracy on the test set across various iterations during the iterative learning. RE denotes the relation extraction in the initial parser. NER denotes the Named Entity Recognition system. Seq2Seq is the model used to detect relations not extracted by the initial parser.}
    \label{fig:curve}
\end{figure}

\paragraph{Iterative Learning}
\autoref{fig:curve} shows the test accuracy versus iterations. Here an iteration is an update on the SMART model based on collected successful samples prior to that iteration, as illustrated in Algorithm~\ref{alg:iterative}. We can see that the model improves during iterative learning and begins to converge at the third iteration.

\paragraph{Ablative Study}
The ablative study in \autoref{fig:curve} analyzes the effect of each module in the inference procedure on the test set. We observe that the sole use of NER only achieves around 20\% accuracy. This indicates that reasoning about relations between attributes is required in most problems. The final result with both the original relation extraction (RE) module and Seq2Seq outperforms the model with just the RE module. This suggests Seq2Seq does detect relations not extracted by RE. 
However, Seq2seq alone is inferior to RE, which suggests explicit mapping of relations works better than implicit learning of a neural module.

\paragraph{Qualitative Study}
To further analyze our model's interpretability and generalization ability, we visualize several examples from the test set, as shown in \autoref{fig:qualitative}.  The first two instances are from the original set, while the last two are from the OOD split.  Both models work well on the first instance since there are similar samples in the training set. When it comes to the third instance, the neural network failed since it has never observed a problem with more than two events and does not comprehend the relations. However, the situation model handles well due to event independence. 
In the fourth instance, the neural network fails to determine the speed of the next car and the relation ``20\% of the total distance", while SMART is good at both.

However, SMART also makes wrong predictions. In the second instance, ``the two vehicles were 35 kilometers apart" is ambiguous since we do not know if the cars have met each other yet. Therefore, 
SMART gives a negative value for the total length. 
In the future, error analysis is needed in the situation model so that when an implausible answer is given, the situation model seeks an alternative solution. 

\section{Conclusions}
In this work, we propose a situation model with attributed grammar for algebra story problems. The experiment results on ASP6.6k indicate our model outperforms state-of-the-art models and shows better interpretability and generalization ability. Future work includes creating an intelligent tutor that helps students develop problem-solving skills.

\newpage

\section{Ethical Impact}
This paper implements the situation model applied by humans to solve algebra story problems on the setting of artificial intelligence. Based on the model, educators can design intelligent tutors to guide students in mathematical learning.

\section{Acknowledgement}
This work reported herein is supported by ARO W911NF1810296, DARPA XAI N66001-17-2-4029, and ONR MURI N00014-16-1-2007. 

\bibliography{reference}
\end{document}